\documentclass[lettersize,journal]{IEEEtran}

\usepackage{amsmath,amsfonts}
\usepackage{algorithmic}
\usepackage{algorithm}
\usepackage{array}
\usepackage[caption=false,font=normalsize,labelfont=sf,textfont=sf]{subfig}
\usepackage{textcomp}
\usepackage{stfloats}
\usepackage{url}
\usepackage{verbatim}
\usepackage{graphicx}
\usepackage{cite}
\usepackage[colorlinks=true, urlcolor=blue, linkcolor=red]{hyperref}
\usepackage{multirow}
\usepackage{multicol}
\usepackage{caption}
\usepackage{subcaption}

\begin{document}

\bibliographystyle{IEEEtranS}

\title{\texttt{LiGuard}: A Streamlined Open-Source Framework for Rapid \& Interactive Lidar Research}

\author{Muhammad Shahbaz and Shaurya Agarwal (Senior Member, IEEE)
\thanks{Muhammad Shahbaz (Ph.D. Student) and Shaurya Agarwal (Associate Professor) are with the Department of Civil, Environmental and Construction Engineering, University of Central Florida, Orlando, FL, USA (emails: muhammad.shahbaz@ucf.edu, shaurya.agarwal@ucf.edu).}%
\thanks{Manuscript received December 12, 2024; revised December 12, 2024.}}

\markboth{Journal of \LaTeX\ Class Files,~Vol.~14, No.~8, August~2021}%
{Shell \MakeLowercase{\textit{et al.}}: A Sample Article Using IEEEtran.cls for IEEE Journals}

\IEEEpubid{0000--0000/00\$00.00~\copyright~2024 IEEE}

\maketitle

\begin{abstract}
There is a growing interest in the development of lidar-based autonomous mobility and Intelligent Transportation Systems (ITS). To operate and research on lidar data, researchers often develop code specific to application niche. This approach leads to duplication of efforts across studies that, in many cases, share multiple methodological steps such as data input/output (I/O), pre/post processing, and common algorithms in multi-stage solutions. Moreover, slight changes in data, algorithms, and/or research focus may force major revisions in the code. To address these challenges, we present \texttt{LiGuard}, an open-source software framework that allows researchers to: 1) rapidly develop code for their lidar-based projects by providing built-in support for data I/O, pre/post processing, and commonly used algorithms, 2) interactively add/remove/reorder custom algorithms and adjust their parameters, and 3) visualize results for classification, detection, segmentation, and tracking tasks. Moreover, because it creates all the code files in structured directories, it allows easy sharing of entire projects or even the individual components to be reused by other researchers. The effectiveness of \texttt{LiGuard} is demonstrated via case studies.
\end{abstract}


\section{INTRODUCTION}
The light detection and ranging (lidar) technology has the capability to scan environments in 3D at an unparalleled accuracy. Until recently, the lidar sensors had low reliability and high cost but due to advances, including innovations in solid-state lidar and nanophotonics-based devices, the technology has improved significantly in terms of performance, reliability, and cost-effectiveness \cite{solid_and_nano_lidar}. Furthermore, the integration of advanced optics and beam steering technologies have improved the spatial resolution and operational efficiency of lidar sensors, facilitating their deployment in large-scale environments \cite{beam_stearing}. These improvements have sparked growing interest in lidar as an active, accurate, and light-agnostic perception technology for intelligent transportation systems (ITS), including applications in autonomous driving, traffic monitoring, and road safety \cite{sun2022object, zimmer2023infradet3d, zhao2019detection}. However, lidar data is inherently complex, posing challenges for researchers in conducting their studies. For instance, in most cases of ITS, a single complete scan from a lidar sensor, also called as point cloud, can have hundreds of thousands of 3D points, leading to substantial computational and data management demands \cite{pcd_data_management_and_processing}. To conduct research on such complex data, it is, at first, necessary to develop robust software tools that can handle and process it efficiently.

Currently, in most cases, to perform an experiment on lidar data, researchers develop code (in computer programming language) tailored to the specific needs of the experiment. This code performs a series of functional steps, hereafter referred to as experiment pipeline, on the lidar (and accompanying data) to conduct the experiment. These steps may include reading data, pre-processing it, running the algorithm/model under research, and (often post-processing the outputs for) getting the results that are evaluated later against some metric and/or visualized for qualitative analysis. The entire code for an experiment pipeline usually spans over multiple interdependent files. This approach to experimenting is common, however, it creates tightly coupled code, posing challenges if a change is needed in the experiment pipeline or the experiment pipeline needs to be reused in some other experiment(s). For example, a researcher may need to run the same experiment on a different dataset with a different file type for point cloud data, disable a specific algorithm in the experiment pipeline for an ablation study, fine-tune certain parameters, or reuse steps from the experiment pipeline that are generally found to be ideal in another unrelated experiment, and so on. These challenges necessitate development of a general-purpose software framework for operating on lidar data to accommodate the diverse needs of the researchers in ITS under a single and easy-to-use interface.

\IEEEpubidadjcol 

\begin{figure*}[h!t!]
    \includegraphics[width=0.98\textwidth]{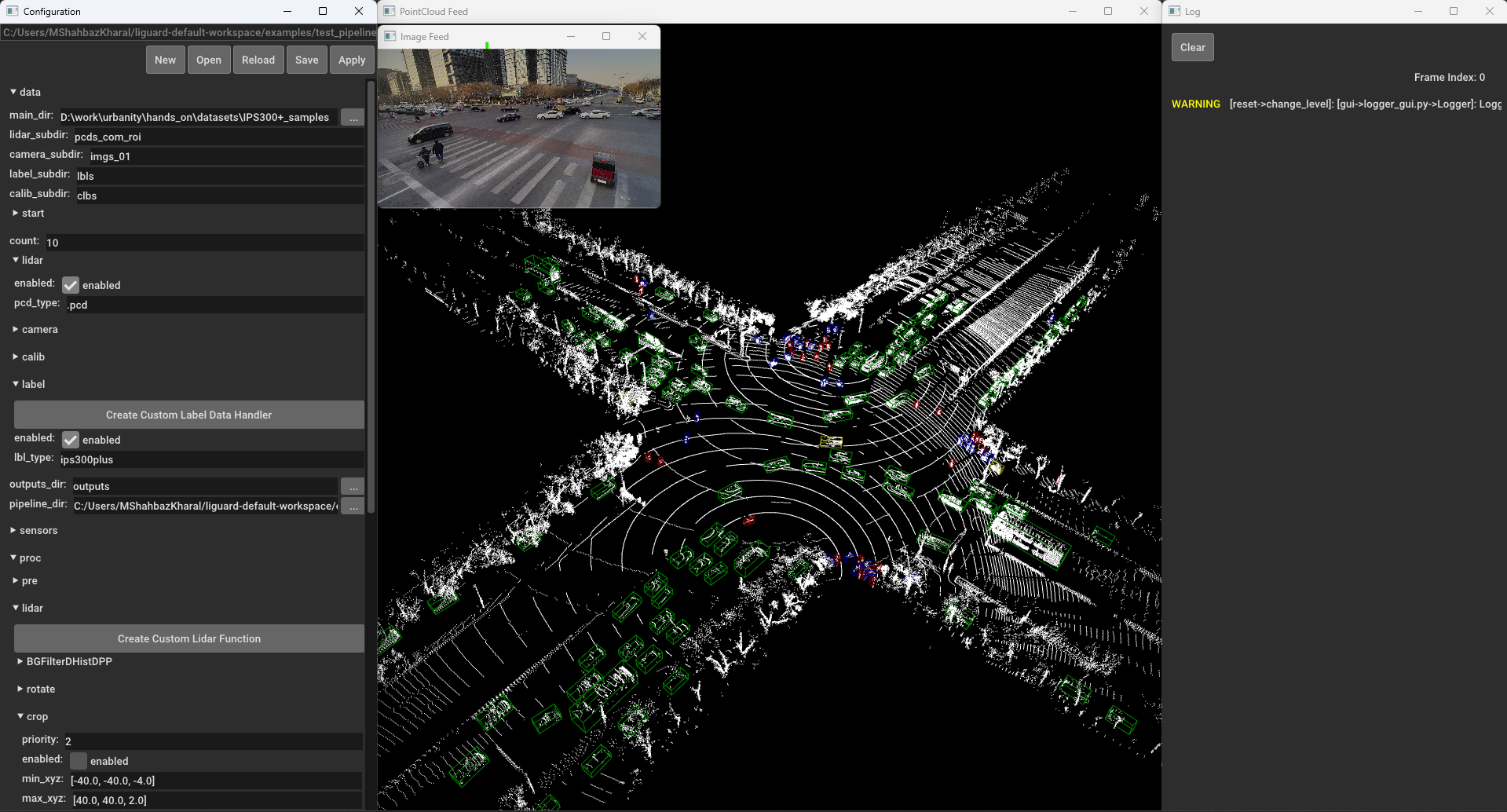}
    \caption{\textbf{\texttt{LiGuard}'s Interface}. \textbf{Left:} The \texttt{Configuration} window allows users to interactively control experiment pipeline; users can leverage built-in data loading capabilities, create, enable/disable/reorder functional steps, dynamically change exposed parameters, and customize visualizations. \textbf{Middle Top:} The \texttt{Image Feed} window visualizes image data. \textbf{Middle Bottom:} The \texttt{PointCloud Feed} window visualizes the lidar data. Both visualizers support common annotation types including bounding boxes, point/pixel colors, and trajectory trails. This figure shows a frame (unprocessed) from IPS300+ \cite{ips300+} dataset. \textbf{Right:} The \texttt{Log} window displays messages (warnings, errors, etc.); useful for debugging experiments.}
    \label{figure:liguard_interface}
\end{figure*}

In response to the said requisite, we present \textbf{\texttt{LiGuard}}. \texttt{LiGuard} is an open-source GUI-based software framework (Figure \ref{figure:liguard_interface}) designed to facilitate rapid, interactive, reusable, and reproducible research on point cloud (and corresponding image) data. \textbf{First}, for a rapid start, it provides numerous built-in features for data loading, processing, and visualization. For data loading, it supports common lidar datasets, an expanding list of lidar and camera sensors, and to some extent, the simulation software CARLA \cite{carla}. For data processing, it offers a range of functions, from basic data pre-processing, such as cropping the input point cloud to a specific region of interest, to advanced yet commonly used algorithms, such as background filtering and point clustering. For visualization, detection (colored bounding boxes), segmentation (colored point clusters), and tracking (line trails) are supported. \textbf{Second}, for an interactive experience, it features a simple graphical-user-interface (GUI). Through GUI, users can interactively enable, disable, and/or reorder functions in their experiment pipelines. These functions can have exposed editable parameters in the GUI making it easy to adjust them and quickly see results in a live and interactive manner. This is different compared to conventional experiment experience where such enabling, disabling, or modifications typically require the researcher to carefully edit code, making sure that it doesn't break other parts of the project, followed by re-executing it. \textbf{Third}, for re-usability, it offers one-click generation of code templates for creating custom data-reader or process functions. The code templates not only offer a standard function structure that directly integrates into \texttt{LiGuard} but also encourage users to create reusable parts in their experiment pipelines that can later be shared across other experiments. \textbf{Forth}, for reproducibility, it manages experiment pipelines in structured directories, simplifying the process of sharing both entire pipelines and individual functions utilized within those pipelines. In summary, \texttt{LiGuard} has following \textbf{key features}:

\begin{itemize}
    \item It supports format-agnostic live and offline data reading.
    \item It offers, built-in, a range of commonly used functions and algorithms for lidar and image data processing.
    \item It has visualizers for both lidar and image data.
    \item It provides GUI for interactivity during experimentation.
    \item It supports the integration of custom research algorithms.
\end{itemize}

\texttt{LiGuard} is designed to be an expandable open-source project encouraging contributors to, over time, add more reusable pipeline components. The project and its technical usage instructions are available at our GitHub repository (\url{https://github.com/m-shahbaz-kharal/LiGuard-2.x}) and a video tutorial series is available at our Youtube channel (\url{https://youtube.com/playlist?list=PLF2HAlDp6JM8_Xsyv0pzJLypH74cradjd}).

\section{Existing Tools for Lidar Data Processing}\label{related_work}

\begin{table*}[ht!]
    \centering
    \begin{tabular}[0.98\textwidth]{|l|c|c|c|c|c|c|c|c|c|c|c|}
        \hline
        \multirow{4}{*}{\textbf{Tools}} & \multicolumn{11}{c|}{Features} \\
        \cline{2-12}
        & Python & Single & Multi- & Point- & Image & Interactive & Offline & Live & Custom & Data- & Visuali-\\
        & Language & Frame & Frame & Cloud & (RGB) & GUI & Data & Sensor & Ops& Type & zation \\
        & Support & Support & Support & Ops & Ops & Support & Support & Support & Support & Agnostic & Support \\
        \hline
        \hline
        \multicolumn{12}{|l|}{\textbf{Libraries}} \\
        \hline
        \hline
        PCL \cite{PCL} & & \checkmark & & \checkmark & \checkmark &  & \checkmark & \checkmark  &  & \checkmark &\checkmark \\
        \hline
        Python-PCL \cite{python-pcl} & \checkmark & \checkmark & & \checkmark & \checkmark &  & \checkmark & \checkmark  &  & & \checkmark\\
        \hline
        PCL.py \cite{pcldotpy} & \checkmark & \checkmark & & \checkmark & \checkmark &  & \checkmark & \checkmark  &  & &\checkmark \\
        \hline
        PCLPy \cite{pclpy} & \checkmark & \checkmark & & \checkmark &  &  & \checkmark &  &  & &  \\
        \hline
        PyPCD \cite{pypcd} & \checkmark & \checkmark & & \checkmark &  &  & \checkmark &  &  &  & \\
        \hline
        PyntCloud \cite{pyntcloud} & \checkmark & \checkmark & & \checkmark &  &  & \checkmark &  &  & \checkmark & \checkmark\\
        \hline
        PyVista \cite{pyvista} & \checkmark & \checkmark & & \checkmark & \checkmark & \footnote[1] & \checkmark &  & \footnote[2] & \checkmark & \checkmark \\
        \hline
        Open3D \cite{Open3D} & \checkmark & \checkmark & & \checkmark & \checkmark & \footnote[1] & \checkmark & \checkmark & \footnote[2] & \checkmark & \checkmark \\
        \hline
        PointCloudSet \cite{pointcloudset} & \checkmark & \checkmark & \checkmark & \checkmark & & & \checkmark & & \checkmark & \checkmark &  \\
        \hline
        PDAL \cite{pdal_contributors_2022_2616780} & \checkmark & \checkmark & \checkmark & \checkmark & & & \checkmark & & \checkmark & \checkmark & \\
        \hline
        \hline
        \multicolumn{12}{|l|}{\textbf{Toolkits \& Software}} \\
        \hline
        \hline
        3DTk \cite{nuchter20113dtk} & & \checkmark & & \checkmark & & & \checkmark & & & \checkmark & \checkmark \\
        \hline
        Paraview \cite{paraview} & & \checkmark & \checkmark & \checkmark & \checkmark & \checkmark & \checkmark & \checkmark & \checkmark & \checkmark & \checkmark \\
        \hline
        Meshlab \cite{meshlab} & & \checkmark & & \checkmark & \checkmark & & \checkmark & & & \checkmark & \checkmark \\
        \hline
        \textbf{\texttt{LiGuard}} & \checkmark & \checkmark & \checkmark & \checkmark & \checkmark & \checkmark & \checkmark & \checkmark & \checkmark & \checkmark & \checkmark \\
        \hline
    \end{tabular}
    \caption{Out-of-Box High-Level Features: \texttt{LiGuard} vs. Other Packages}
    \label{table:liguard_comparison_to_others}
\end{table*}
{\footnotetext[1]{Requires user to implement GUI manually.}}
{\footnotetext[2]{Requires the user to register those callbacks manually.}}

Point cloud data varies in range, resolution, field-of-view, point density, and frame rate based on applications. For aerial surveys, large range and field-of-view are critical, while structural analysis requires high resolution in a narrow field-of-view. ITS applications typically need medium range, resolution, good density, and high frame rates. Therefore, this section focuses on data specifications that are applicable to ITS applications. The libraries/software, such as entwine \cite{entwine}, lid-R \cite{lidr}, LASPy \cite{laspy}, and CloudCompare \cite{cloudcompare} focus processing single massive point clouds, are out of the scope for this paper. The review is divided into two sections: libraries and software/toolkits (Table \ref{table:liguard_comparison_to_others}). Libraries are importable code packages, while software/toolkits are complete lidar data processing applications.

\subsection{Point Cloud Processing Libraries}

Point Cloud Library (PCL) \cite{PCL} is, arguably, the most efficient library to operate on point cloud data. However, the core PCL is written in C++ programming language and is therefore challenging to use, especially, if the research project involve rapid changes in the code. In response, a number of Python-bindings for PCL emerged, including Python-PCL \cite{python-pcl}, PCL.py \cite{pcldotpy}, and PCLPy \cite{pclpy}. However, the former two libraries lack most of the original PCL features due to the limitations of Cython \cite{cython}, the binding generator used to port PCL to Python. The latter uses a more robust PyBind11\cite{pybind11} to bind to most of PCL, however, it still lacks visualization and some other modules. Moreover, the code repository of PCLPy has been inactive for the last 3 years. A pure python re-implementation of PCL, PyPCD \cite{pypcd}, is also available, however, according to the official documentation, it is not a production ready library at the time of publication of this paper. Cilantro \cite{cilantro}, another efficient C++ library for operating on point cloud data,
is also challenging to utilize due to the lack of Python bindings.

The libraries that are both actively maintained and support Python programming language include PyntCloud \cite{pyntcloud}, PyVista \cite{pyvista}, and Open3D \cite{Open3D}. However, only the latter two are suitable for multi-frame processing as PyntCloud is designed for operations on singular point clouds. PyVista and Open3D both are feature rich libraries that support operations ranging from basic transformations (e.g., translation, rotation, etc.) and manipulations (e.g., cropping, clustering, etc.) to advanced operations (e.g., registration, surface reconstruction, etc.). \texttt{LiGuard} also uses Open3D heavily under the hood for many of its operations. Nonetheless, it is important to underscore that despite the role of these libraries as bases for development of advance 3D applications, it is challenging to directly use them interactively in research experiments.

For creating experiment pipelines that involve bulk data processing, the most significant efforts are presented by the pointcloudset \cite{pointcloudset} and point data abstraction library (PDAL) \cite{pdal_contributors_2022_2616780}. These can handle complex processing pipelines, support concurrent execution, and allow custom function calling over bulk lidar data, however these are designed to operate on pre-stored point-cloud-only data limiting their use in ITS applications that often require both point cloud and image data.

\subsection{Point Cloud Processing Software \& Toolkits}
The 3DTk toolkit \cite{nuchter20113dtk} offers a suite of point cloud operations (including automatic registration, 6D Simultaneous Localization and Mapping (SLAM), plane extraction, and visualization) through a command-line interface. However, the C++ source code is challenging to modify, hindering the customization required for the research. Furthermore, the command-line interface (CLI) introduces operational complexity. Tools like Paraview \cite{paraview} and Meshlab \cite{meshlab} mitigate the CLI challenges by offering GUI, but the underlying C++ backend still presents obstacles to customization.

\section{\textbf{\texttt{LiGuard}}'s Design}\label{liguard_design}

\begin{figure*}[h!t!]
  \centering
  \includegraphics[width=\textwidth]{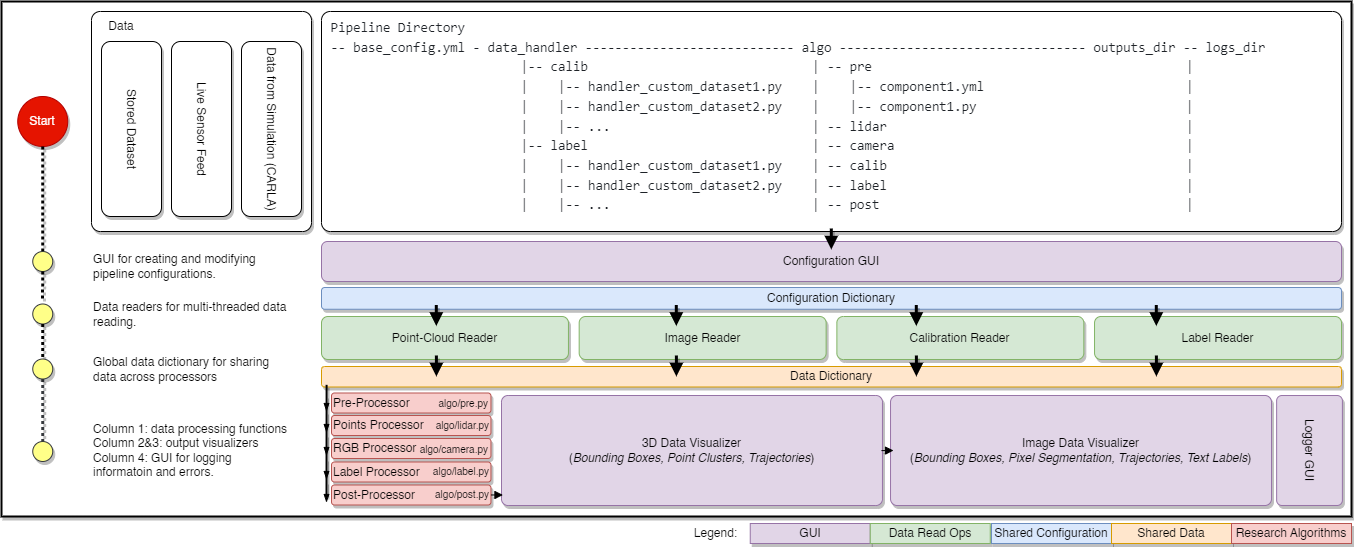}
  \caption{\textbf{\texttt{LiGuard}}'s Architecture}
  \label{figure:liguard_arch}
\end{figure*}

The primary objective of \texttt{LiGuard} is to offer a framework for research that require rapid and interactive adjustments to experiment pipelines involving lidar and image data. It also aims to encourage re-usability and reproducibility of those experiment pipelines. Therefore, it is designed to focus modularity that, in the context of \texttt{LiGuard}, means that every functional step should be independent of all the other functional steps in an experiment pipeline. This does not mean that functions cannot rely on data produced by other functions, as such dependencies are intrinsic to experiment pipelines. Rather, it ensures that no function is hardcoded to invoke another in its implementation. As long as the required data is available in the data stream, a function can perform its task and return the processed data to the pipeline for use by other functions. Such design offers versatility to experiment pipelines as it allows enabling, disabling, modifying, reordering, and even replacing individual steps during execution providing interactivity to experimentation.

\subsection{Architecture}

\texttt{LiGuard} is developed on top of two feature-rich libraries Open3D \cite{Open3D} (for operating on point cloud data and creating graphical-user-interface) and OpenCV \cite{opencv} (for operating on image data),
to offer flexibility in expansion. Its architecture, as presented in the Figure \ref{figure:liguard_arch}, is divided into five major components including 1) graphical user interfaces or GUIs (colored purple) for effortless interactions, 2) shared configuration dictionary (colored blue) for keeping configuration data of the pipeline, 3) individual data readers (colored green) for type-agnostic data parsing of point cloud, image, calibration, and label data, and 4) shared data dictionary (colored yellow) for allowing independent execution of 5) built-in/custom algorithms (colored red).

\subsection{Functionalities}

\texttt{LiGuard} presents, at start, an interactive GUI titled \texttt{Configuration} that manages creation, modification, and running of experiment pipelines. It offers setting directory paths if data is in the form of a stored dataset, or connection parameters if the data source is a live physical sensor or CARLA \cite{carla} (an autonomous driving) simulation software. It also allows, in a live and interactive manner, toggling state (to be enabled or disabled), changing priority (order) of execution, and editing parameters, of the functions in the pipeline.

The built-in functions/algorithms in \texttt{LiGuard} are populated by default in the \texttt{Configuration} window and can be utilized just by enabling them. A list of all built-in functions, in the current release (2.1.3), is presented in Table \ref{table:liguard_functional_features}. To add custom functions (that can be as small as single operation or as complex as  complete algorithm) to the experiment pipeline, the user can utilize on-click \texttt{Create Custom Function} button in the \texttt{Configuration} window. This automatically opens up the default code editor and creates a function template that is standard to \texttt{LiGuard}. The created function has access to all the configuration and the data in the pipeline. The creation or modification of functions does not require \texttt{LiGuard} to be restarted making it easy to edit and test code.

\begin{table}[h!]
\centering
\resizebox{0.98\columnwidth}{!}{%
\begin{tabular}{|l|l|l|}
    \hline
    \textbf{Category}           & \textbf{Algorithm/Function}          & \textbf{Short Description}           \\
    \hline
    \hline
    \multirow{5}{*}{Pre-Processing} & remove\_nan\_inf\_allzero\_from\_pcd & Removes NaN \& Inifity Values        \\
    \cline{2-3}
                                & \multirow{3}{*}{manual\_calibration} & Creates lidar to cam (Tr\_velo\_to\_cam) matrix \\ 
                                \cline{3-3} & & Creates rectification (R0\_rect) matrix \\
                                \cline{3-3} & & Create projection (P2) matrix                  \\ \hline
    \multirow{10}{*}{
            Point Cloud Operations
        } & rotate         & Rotates using euler angles  \\
        \cline{2-3} 
                                & crop                              & Crops to Min \& Max Bounds (x,y,z)     \\ 
        \cline{2-3} 
                                & project\_image\_pixel\_colors                                 & Colorizes using pixel colors and calibration matrices              \\
        \cline{2-3} 
                                & BGFilterDHistDPP                                 & Applies Dynamic Histogram Per Point background filter          \\
        \cline{2-3} 
                                & BGFilterSTDF                                 & Applies Spatio-Temporal Density background filter         \\
        \cline{2-3} 
                                &  Clusterer\_TEPP\_DBSCAN \cite{tepp_db_scan}                                 & Clusters using Theoretically Efficient and Practical Parallel DBSCAN            \\
        \cline{2-3} 
                                & O3D\_DBSCAN                                 & Clusters using DBSCAN implemented in Open3D \cite{Open3D}             \\
        \cline{2-3} 
                                & Cluster2Object                                 & Converts lidar clusters to object labels using size heuristics             \\
        \cline{2-3} 
                                & PointPillarDetection                                 & Detects objects using PointPillars \cite{pointpillars} algorithm           \\
        \cline{2-3} 
                                & gen\_bbox\_2d                                 & Gets 2D bboxes from 3D bboxes using calibration matrices          \\
                                
    \hline
    \multirow{2}{*}{Image Operations}                       & project\_point\_cloud\_points                              & Projects point cloud onto image using calibration matrices             \\
        \cline{2-3} 
                                & UltralyticsYOLOv5                 & Detects objects using Ultralytics YOLOv5 object detection models    \\
    \hline
    \multirow{2}{*}{Label Operations} & remove\_out\_of\_bound\_labels                     & Removes labels that are out of the specified bounding box.    \\ \cline{2-3} 
                                & remove\_less\_point\_labels                            & Remove labels with fewer points than the specified threshold.                    \\ \cline{2-3} 
    \hline
    \multirow{7}{*}{Post-Processing} & Fuse2DPredictedBBoxes & Fuses bounding boxes information among ModalityA and ModalityB. \\ \cline{2-3} 
    \cline{2-3} 
                                & GenerateKDTreePastTrajectory                                 & Generates past trajectory of objects using KDTree matching.          \\
    \cline{2-3} 
                                & GenerateCubicSplineFutureTrajectory                                 & Generates future trajectory using cubic spline interpolation.          \\
    \cline{2-3} 
                                & GeneratePolyFitFutureTrajectory                                 & Generates future trajectory using polynomial fitting.          \\
    \cline{2-3} 
                                & GenerateVelocityFromTrajectory                                 & Generates velocities of objects from their trajectories.          \\
    \cline{2-3} 
                                & create\_per\_object\_pcdet\_dataset                                 & Creates a per-object PCDet dataset from current label data in pipeline.          \\
    \cline{2-3} 
                                & create\_pcdet\_dataset                                & Creates PCDet dataset from current label data in pipeline.          \\
                                
    \hline
    \end{tabular}%
    }
    \caption{Built-in Functions/Algorithms in \textbf{\texttt{LiGuard}}}
    \label{table:liguard_functional_features}
\end{table}

The application logic of \texttt{LiGuard}, that is, how data and configuration are read, shared across pipeline processes, and visualized, is decoupled from functional steps of the pipeline, facilitating users to focus on research rather than the intricacies of developing high-quality code.

\subsection{Data Flow \& Operation}

The experiment pipelines in \texttt{LiGuard} are organized into structured directories (see \textit{Pipeline Directory} in Figure \ref{figure:liguard_arch}). Users can create and initialize a new pipeline directory using the \texttt{New} button in the \texttt{Configuration} window. The primary configuration file in the created directory, \texttt{base\_config.yml}, contains commonly used settings including data directories, visualization options, and debugging parameters, etc. It also includes configurations for built-in features under the section \texttt{proc} (short for processes). Each sub-section under \texttt{proc} represents a process category. There are six process categories that are currently supported, named 1) \texttt{pre}, 2) \texttt{lidar}, 3) \texttt{camera}, 4) \texttt{calib}, 5) \texttt{label}, and 6) \texttt{post}. The processes under \texttt{pre} category, as the name suggests, are responsible for pre-processing the data. Categories 2-5 operate on point cloud, image, calibration, and label data, respectively. The \texttt{post} process category handles the post-processing operations on the data. There can be multiple functions under each process category and their order of execution depends upon \texttt{priority}, a live-editable number (lower is higher). Any other parameters defined under each function are also exposed in the GUI and can be live-edited during execution of the pipeline.

Users can add custom functional steps to their experiment pipelines by clicking \texttt{Create Custom Function} button in \texttt{Configuration} window under \texttt{proc} section. It opens up a dialog asking function's name, followed by creation of code (.py) and configuration (.yml) files under \textit{\textless pipeline\_directory\textgreater /algo/\textless process\_category\textgreater/} path. These files are automatically opened in the default text editor and contain template along detailed comments to help user quickly start developing the required functional step. Similarly, users can create custom data loaders by clicking \texttt{Create Custom Data Handler} button under \texttt{data} section. An in-depth look into \texttt{LiGuard} pipelines is available at our Github Documentation Pages (\url{https://m-shahbaz-kharal.github.io/LiGuard-2.x/liguard_pipelines.html}).

    


\begin{figure*}[ht!]
  \centering
  \includegraphics[width=\textwidth]{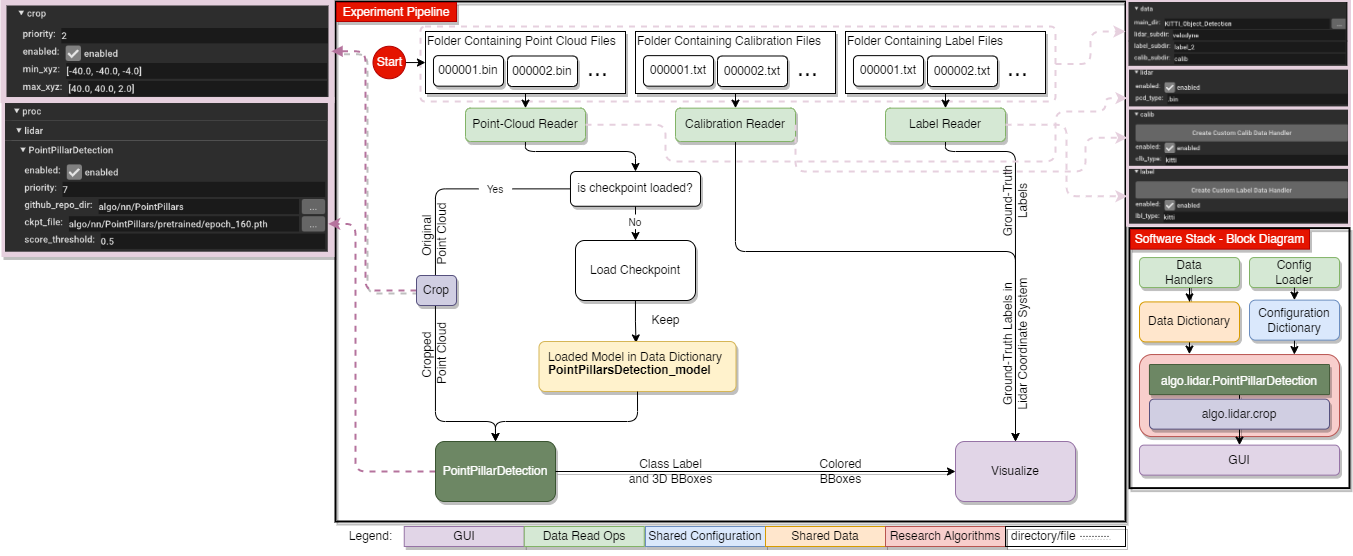}
  \caption{\textbf{Pipeline 1}: PointPillars \cite{pointpillars} Inference on KITTI \cite{kitti} Dataset. In the \textbf{Center} the flow of the experiment pipeline is shown that is color coded according to \textbf{Right Bottom} Software Stack diagram. On \textbf{Left \& Right Top}, in pink bordered boxes, the corresponding GUI components are shown to visualize the pipeline relation to \texttt{LiGuard}'s \texttt{Configuration} window.}
  \label{figure:pipeline_1}
\end{figure*}

\section{Technology Demonstration}\label{technology_demonstration}

The application of lidar technology in ITS research spans a diverse range of focuses, reflecting the breadth of challenges and interests in the field. Consequently, it is important to present \texttt{LiGuard} in a manner that underscores its utility and flexibility. This section introduces two distinct experiment pipelines as case studies that are also included as examples in standard \texttt{LiGuard} installation for reproducibility purpose. The primary objective of these case studies is to demonstrate how \texttt{LiGuard} facilitates a rapid start, promotes component reuse, and supports fine-tuning of functional steps in experiment pipelines. However, to highlight the flexiblity of \texttt{LiGuard}, these pipelines, in order, increase in complexity and are intentionally unrelated to each other. Through this approach, we establish the  general applicability of \texttt{LiGuard} that not only reduces the redundancy inherent in conventional research methods but also promotes efficiency and adaptability across diverse ITS applications.

\subsection{Case Study 1: Inference of Pre-Trained Deep Object Detector on Pre-Recorded Dataset}
In this case study, a straightforward use of \texttt{LiGuard} is demonstrated. The experiment pipeline involves three steps: loading point cloud data from a standard dataset, passing it through a pre-trained deep object detector to get bounding box predictions, visually compare the predictions and ground-truth labels. This example highlights how researchers can rapidly set up experiments using built-in components in \texttt{LiGuard}. The dataset being used is KITTI Object Detection Dataset \cite{kitti} and the deep object detector is PointPillars \cite{pointpillars} that is pre-trained on the KITTI dataset.

Creating an experiment pipeline around PointPillars needs some pre and post processing steps. In pre-processing, the point cloud is cropped to region of interest. Also, the output of the model is a list of labels in lidar coordinate system but the ground-truth labels in KITTI dataset are in camera coordinate system. To compare them, some post-processing is needed that involves converting coordinate system from camera to lidar space using calibration data.

\texttt{LiGuard} has built-in support for reading KITTI dataset, pre-processing, post transformations of the coordinate systems, and visualizations, so a re-implementation of them is not needed. In conventional setup, all of these steps are usually repeated across experiments but in this case the only development needed is loading the model, passing it the input, getting the predictions, and adding them back to shared data dictionary. The complete pipeline (Figure \ref{figure:pipeline_1}) can be defined, step-by-step, as follows:

\vspace{6pt}
\hrule
\vspace{2pt}
\noindent\textbf{Pipeline 1: Deep Object Detector Inference on Dataset}
\vspace{2pt}
\hrule
\vspace{4pt}
\begin{enumerate}
    \item Read point cloud, calibration, and (ground-truth) label data from KITTI dataset.
    \item Crop the point cloud to region of interest.
    \item If not loaded, load PointPillars pre-trained model.
    \item Pass the point cloud data through PointPillars, get bounding box detections, and put them back in the shared data dictionary.
    \item Use calibration data to convert ground-truth bounding boxes to lidar coordinate system.
    \item Visualize predicted and ground-truth bounding boxes.
\end{enumerate}
\vspace{4pt}
\hrule
\vspace{6pt}

\begin{figure}[h!]
\centering
\begin{tabular}{cc}
    \includegraphics[width=0.44\linewidth]{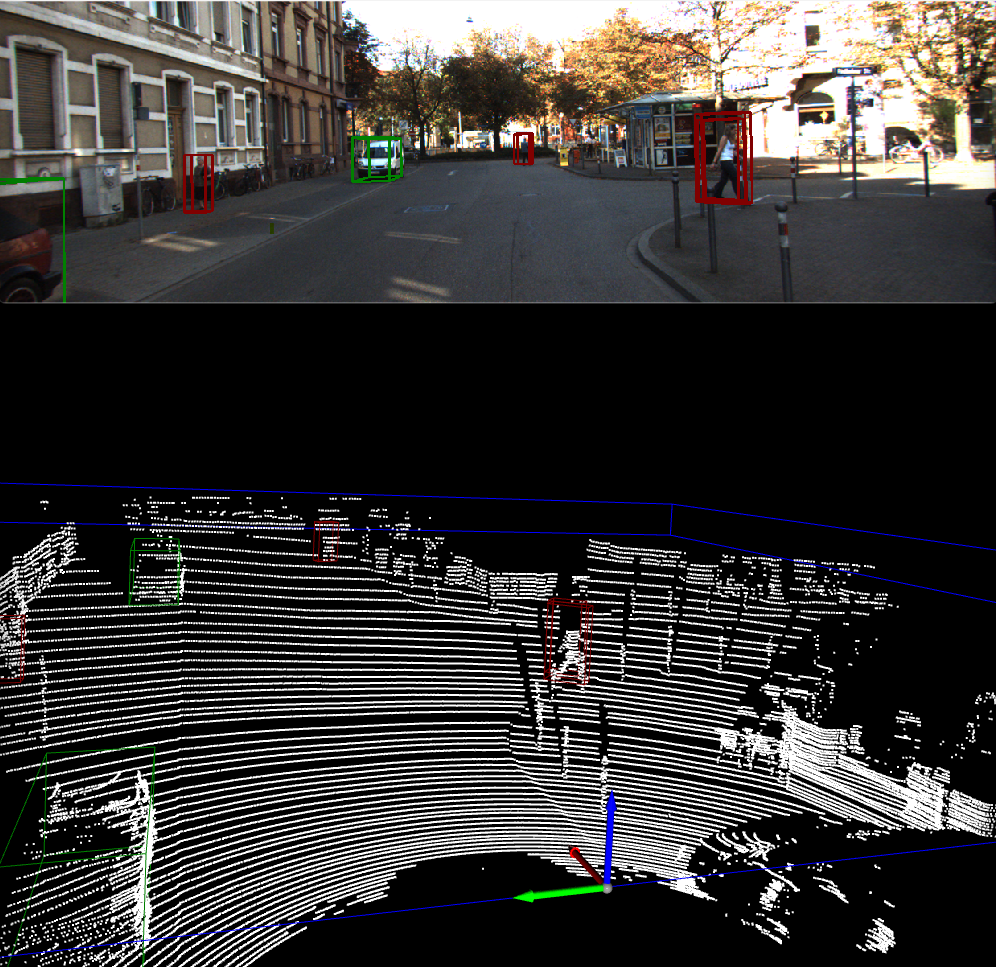} &
    \includegraphics[width=0.44\linewidth]{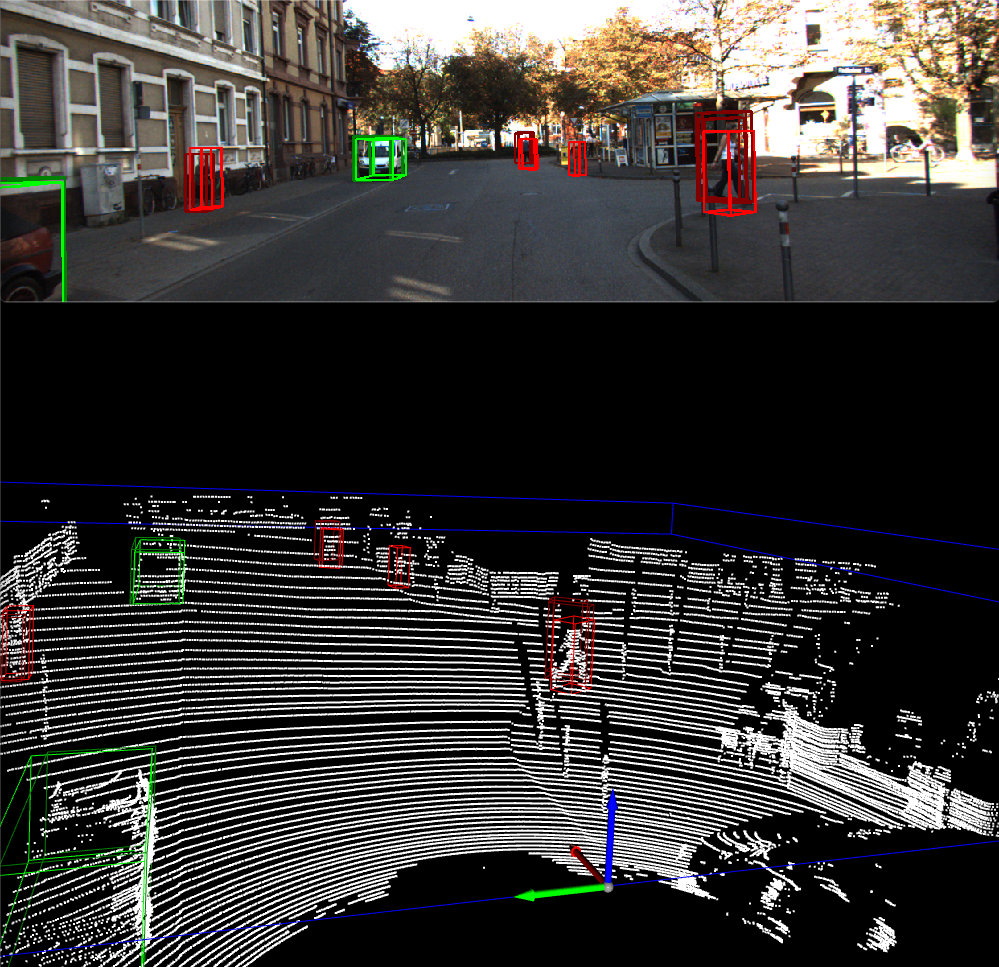}
\end{tabular}
\caption{A point cloud frame from KITTI \cite{kitti} dataset passed through Pipeline 1. The sub-figures are crops from the \texttt{Image Feed} and \texttt{PointCloud Feed} windows of \texttt{LiGuard}'s interface shown in Figure \ref{figure:liguard_interface}. \textbf{Left}: the original point cloud and ground-truth bounding boxes (light colored). \textbf{Right}, the ground-truth and predictions (dark colored).}
\label{figure:pointpillars_on_kitti}
\end{figure}

\noindent For this experiment pipeline, the step 1,2,5, and 6 are  pre-supported by \texttt{LiGuard}. For step 3 and 4, a custom lidar function that implements loading and inferencing the model needs to be created by clicking \texttt{Create Custom Lidar Function} in the \texttt{Configuration} window. For convenience purpose, this experiment pipeline (including the custom function that implements PointPillars utilizing a Github repository \cite{pointpillars_implementation}) is provided in the examples folder. The visualization of the stages before and after applying the PointPillar algorithm is shown in Figure \ref{figure:pointpillars_on_kitti}.

\begin{figure}[h!]
\centering
\begin{tabular}{cc}
    \includegraphics[width=0.44\linewidth]{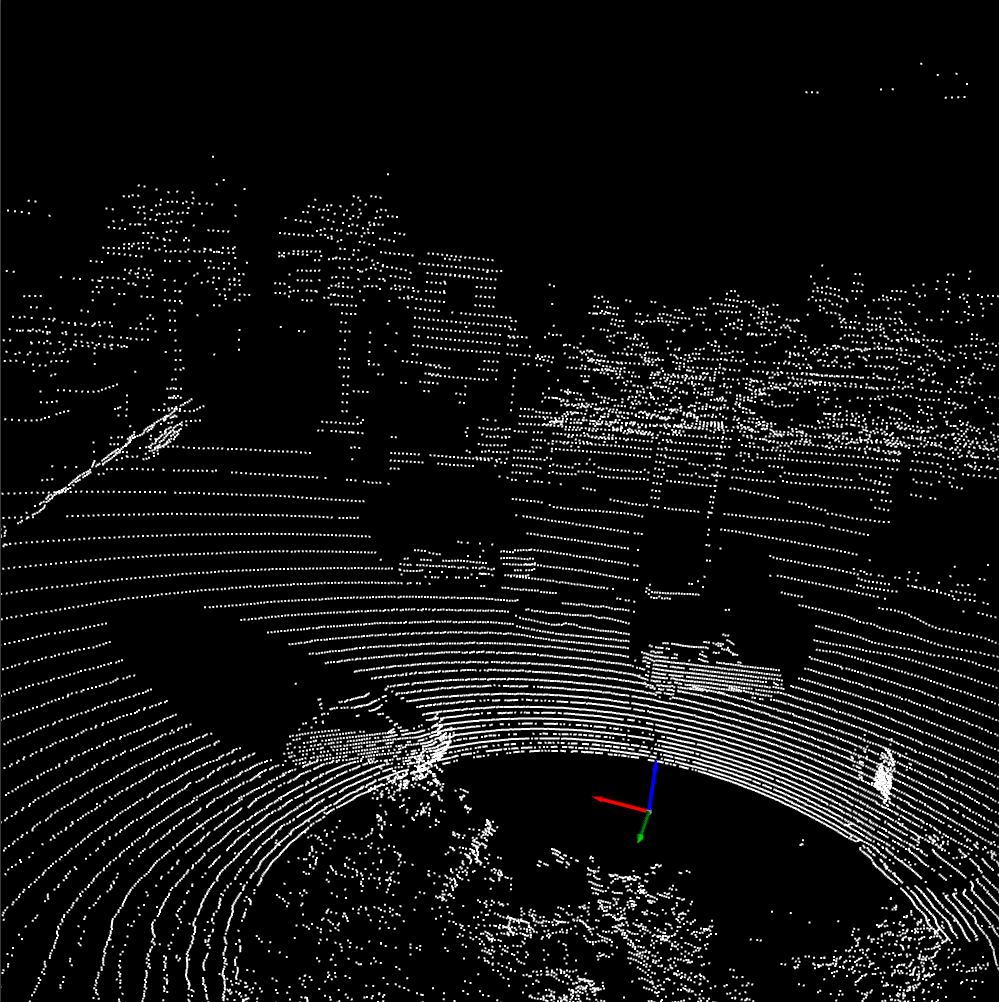} &
    \includegraphics[width=0.44\linewidth]{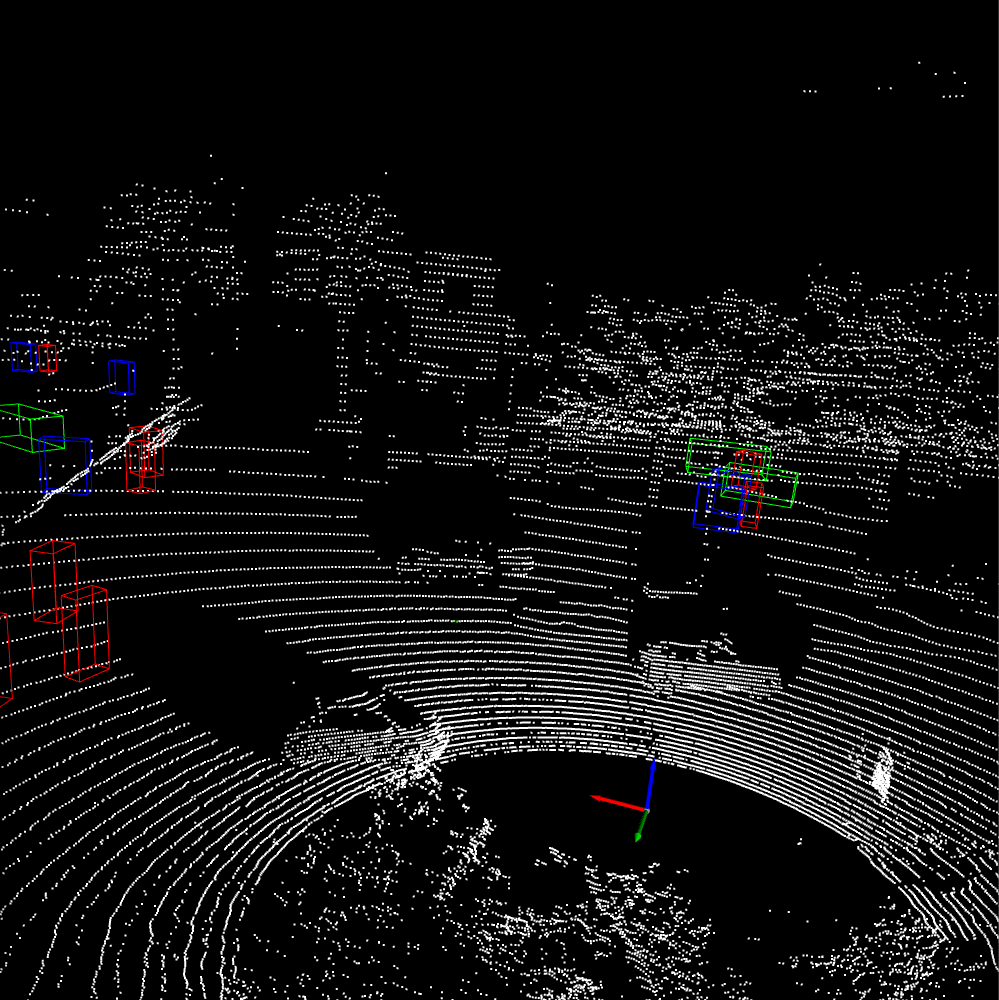}
\end{tabular}
\caption{A point cloud frame from a custom dataset passed through Pipeline 1. \textbf{Left} is the original point cloud. \textbf{Right}, the predictions (dark colored). The model is clearly hallucinating as the data is completely different.}
\label{figure:pointpillars_on_custom_dataset}
\end{figure}

To present the re-usability of \texttt{LiGuard}'s pipelines, let's use a custom dataset instead of KITTI in the same pipeline. The custom dataset is provided under examples folder in the standard installation of \texttt{LiGuard}. The data is recorded using Ouster OS1-64 \cite{ouster_os1_64} lidar sensor from a roadside perspective. To save space, it consists of only 10 point cloud (.pcd) files. To use this dataset in the Pipeline 1, user just need change the \texttt{main\_dir} and \texttt{lidar\_dir} directories to point to the example dataset, disable reading calibration and label data, as labels for this dataset are not provided, by unchecking the checkbox for \texttt{calib} and \texttt{label} under \texttt{data} drop-down, change the \texttt{pcd\_type} to .pcd, all in \texttt{Configuration} window without changing anything in the code. The output of this pipeline is shown in Figure \ref{figure:pointpillars_on_custom_dataset}.

\subsection{Case Study 2: Complex Pipeline for Generating Self-Supervised Labels for Custom Roadside Lidar Dataset}

\begin{figure*}[ht!]
  \centering
  \includegraphics[width=\textwidth]{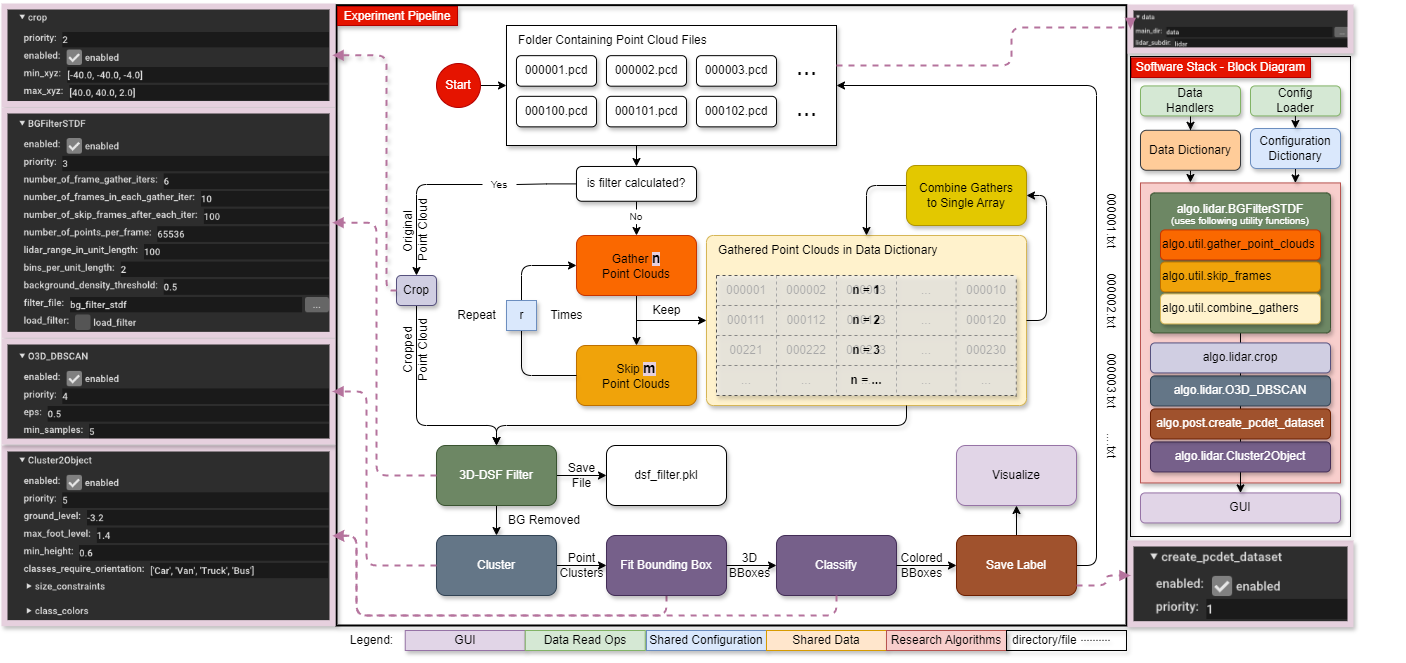}
  \caption{\textbf{Pipeline 2}: Self-Supervised Labeling for Custom Dataset. In the \textbf{Center} the flow of the experiment pipeline is shown that is color coded according to \textbf{Right Center} Software Stack diagram. On \textbf{Left} and \textbf{Right} (Top and Bottom), in pink bordered boxes, the corresponding GUI components are shown to visualize the relation to \texttt{LiGuard}'s \texttt{Configuration} window.}
  \label{figure:pipeline_01}
\end{figure*}

\begin{figure*}[ht!]
\centering
\begin{tabular}{cccc}
    \includegraphics[width=0.22\linewidth]{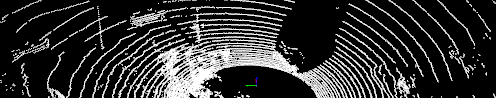} &
    \includegraphics[width=0.22\linewidth]{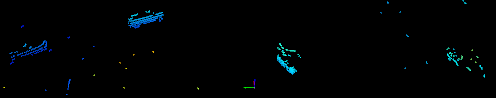} &
    \includegraphics[width=0.22\linewidth]{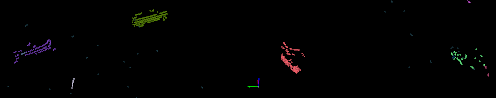} &
    \includegraphics[width=0.22\linewidth]{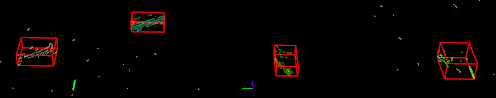} \\
    original point cloud & background removed & clustered & bbox detections
\end{tabular}
\caption{\textbf{Pipeline 2 Stages}. The sub-figures, from left to right, display original point cloud followed by the processed output from \texttt{BGFilterSTDF}, \texttt{O3D\_DBSCAN}, and \texttt{Cluster2Object} functions, respectively.}
\label{figure:self_supervised_on_custom_dataset}
\end{figure*}

\begin{figure*}[hb!]
  \centering
  \includegraphics[width=\textwidth]{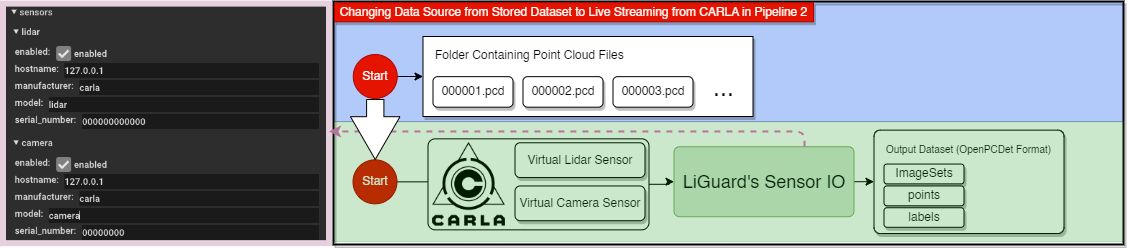}
  \caption{\textbf{Pipeline 2-Modified}. To modify pipeline 2 to use CARLA \cite{carla} as data source instead of a stored dataset, the only required change is disabling everything under \texttt{data} and enabling \texttt{lidar} under \texttt{sensors} in the \texttt{Configuration} window. The rest of pipeline stays exactly the same, and is therefore omitted from this figure.}
  \label{figure:pipeline_03}
\end{figure*}

In this case study, a complex multi-stage experiment pipeline is explored to demonstrate the usefulness of \texttt{LiGuard}. Here, object detection is studied in point cloud data that is recorded using a non-moving roadside lidar sensor. Such problems has recently gained recognition for various road safety applications in the ITS domain \cite{zhao2019detection}, \cite{wu2018automatic}, \cite{sun20183}. To make the case study more illustrative of \texttt{LiGuard}'s use cases, the pipeline is carefully devised to be a multi-stage, highly tunable experiment. It has 6 stages, 1) a number of point clouds are gathered and a spatio-temporal density background filter is computed to remove background points, 2) the remaining points are clustered using DBSCAN algorithm, 3) oriented bounding boxes are regressed using minimum bounding box algorithm in Open3D \cite{Open3D}, 4) bounding boxes are classified into road-user types (cars, pedestrians, etc.) using size heuristics, 5) the labels are stored in a format compatible with OpenPCDet \cite{openpcdet}, a tool-box for training deep object detectors, and 6) bounding boxes are visualized for qualitative assurance. There are many tunable parameters in stage 1,2,3, and 4 making it a highly iterative experimentation scenario that is challenging to tackle using conventional coding. However, due to interactive nature of \texttt{LiGuard}, the parameters can be tuned via GUI and visualized instantly to get robust fine-tuning. The complete pipeline (Figure \ref{figure:pipeline_01}) is defined as follows:

\vspace{6pt}
\hrule
\vspace{2pt}
\noindent\textbf{Pipeline 2: Generating Labels for Roadside Point Cloud Data Using Multi-Stage Self-Supervised Method}
\vspace{2pt}
\hrule
\vspace{4pt}
\begin{enumerate}
    \item Gather chronologically distant frames from the roadside lidar dataset. The chronologically distant frames are important to avoid locations of momentarily stopped vehicles, pedestrians, etc., getting marked as background locations. This can be done by doing $r$ iterations of the following sub-steps
        \begin{enumerate}
            \item Gather $n$ point cloud frames/files.
            \item Skip $m$ point cloud frames/files.
        \end{enumerate}
    \item Calculate a spatio-temporal density-based background filter using some threshold.
    \item Then, for every frame, perform the following sub-steps:
        \begin{enumerate}
            \item Crop the point cloud to the region of interest.
            \item Using calculated filter, remove background points.
            \item Cluster the remaining (foreground) points.
            \item Fit an oriented bounding box on every cluster.
            \item Save label in OpenPCDet \cite{openpcdet} standard format.
            \item Visualize the generated bounding boxes.
        \end{enumerate}
\end{enumerate}
\vspace{4pt}
\hrule
\vspace{6pt}

For this pipeline, each functional step (background filtering, clustering, bounding box regression, classification, and storing data) is developed using \texttt{LiGuard}'s standard function template created using \texttt{Create Custom Function} button in the GUI. However, since these functions are generally useful, they are now included in \texttt{LiGuard}'s built-in reusable component list. The \texttt{crop} function is for cropping the point cloud to the region of interest, \texttt{BGFilterSTDF} for gathering chronologically distant frames, calculating background filter, and using the computed filter to keep only the foreground points, \texttt{O3D\_DBSCAN} for clustering, and \texttt{Cluster2Object} for fitting bounding boxes and assigning classification label. The post-processing operation, \texttt{create\_per\_object\_pcdet\_dataset} is for storing the labels in OpenPCDet \cite{openpcdet} standard format. The dataset used in this pipeline can be downloaded from our Google Drive at \url{https://drive.google.com/drive/folders/1Mk0rzaWyH7-klrVvFnNIxz5IQ7-IOztv?usp=sharing}. The visualization of the stages of this pipeline are shown in Figure \ref{figure:self_supervised_on_custom_dataset}. As each function in this experiment pipeline is a standalone step designed using \texttt{LiGuard}'s standard function template, it can be enabled/disabled, reordered, and even reused in other pipelines. Each function has editable parameters that can be tuned in live and interactive manner allowing users to quickly adjust and see the results.

Recently, we have added support for CARLA \cite{carla} as data source to \texttt{LiGuard}. The above pipeline can be directly be used on data live streaming from CARLA. The user only need to enable \texttt{lidar} under \texttt{sensors} section in \texttt{Configuration} window and set \texttt{hostname} and \texttt{manufacturer} to "carla". The only difference, essentially, is difference in data source (Figure \ref{figure:pipeline_03}).

\section{Conclusion \& Future Work}\label{conclusion_and_future_work}
This paper presented \texttt{LiGuard}, a framework designed to ease the research on lidar and accompanying data for ITS applications. Its key contributions include an interactive GUI-based interface, built-in support for common pipeline components, ease of adding custom functionality, and experiment sharing and customizations. By abstracting the data interfacing and pipeline execution, \texttt{LiGuard} empowers ITS researchers to focus more on the research code rather than application logic code. Its effectiveness is highlighted through illustrative case studies.

\texttt{LiGuard} promises a practical tool for rapid development of point cloud research experiments, however, there are several areas of improvement, including broadening the built-in component library, increasing support for more publicly available datasets, improving its execution performance, increasing technical support by providing more examples, and community engagement. An active effort is being put into making all these improvements in the hope that open-source contributions will aid the speed of overall advancements in the framework.

\bibliography{bib}

\newpage

\section{Biography Section}
 
\vspace{11pt}

\begin{IEEEbiography}[{\includegraphics[width=1in,height=1.25in,clip,keepaspectratio]{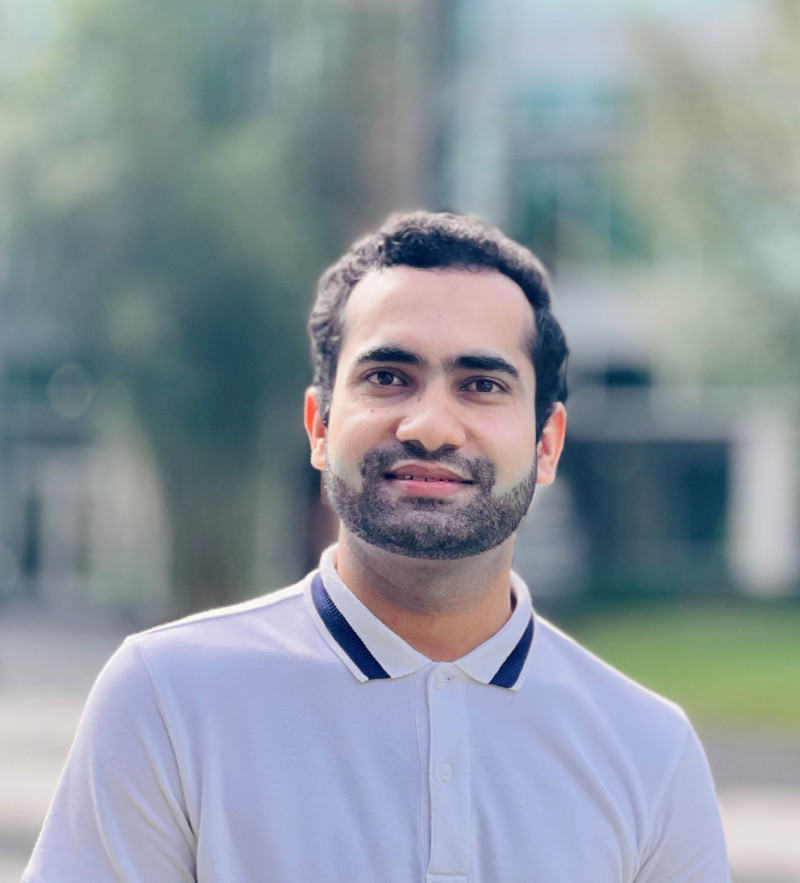}}]{Muhammad Shahbaz} is last year Civil Engineering Ph.D. student at University of Central Florida (UCF). He joined UCF in 2021 as graduate research assistant. His Bachelor's and Master's degrees are in Computer Science. His current research focuses Deep Learning for 3D Computer Vision and Multi-Modal Perception Systems. During his Ph.D. he took major part in multiple funded projects from Florida Department of Transportation (FDOT) for pedestrian safety at intersections and smart work zones. \\
Phone: +1 (321) 200-0125 \\
Email: \url{m.shahbaz.kharal@outlook.com} \\
Github: \url{github.com/m-shahbaz-kharal}. \\
\end{IEEEbiography}

\vspace{11pt}

\begin{IEEEbiography}[{\includegraphics[width=1in,height=1.25in,clip,keepaspectratio]{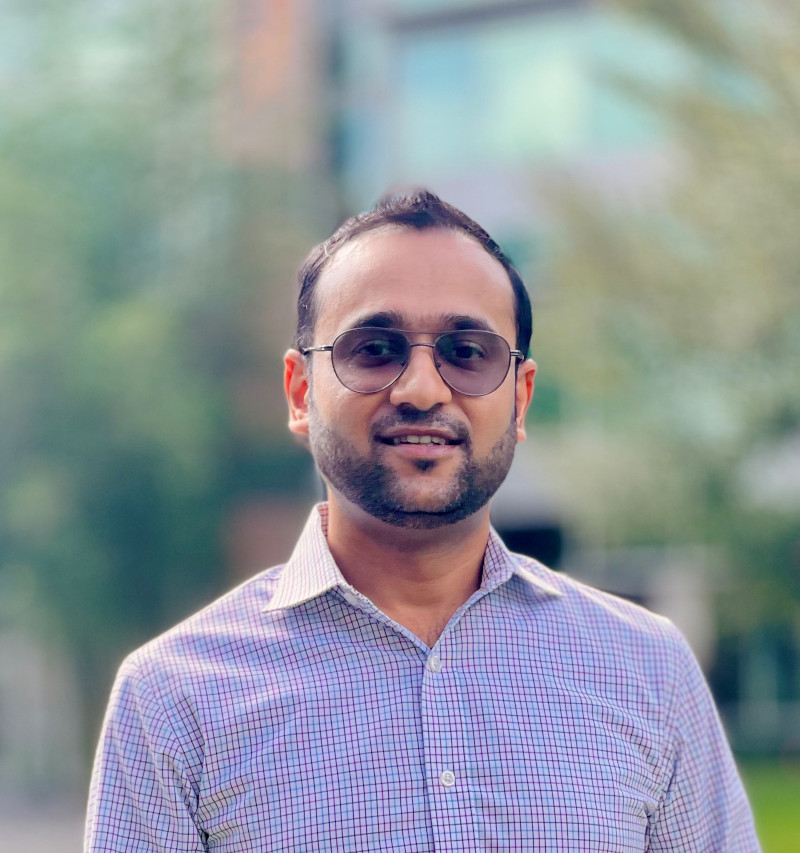}}]{Shaurya Agarwal}
Shaurya Agarwal is currently (2024-present) an Associate Professor in the Civil, Environmental, and Construction Engineering Department at the University of Central Florida. He joined UCF in 2018 as an assistant professor and key member of the Future City Initiative. He is the founding director of the Urban Intelligence and Smart City (URBANITY) Lab and currently serves as the director of Future City Initiative. His research focuses on interdisciplinary areas of cyber-physical systems, smart and connected transportation, and connected and autonomous vehicles. Passionate about cross-disciplinary research, he integrates control theory, information science, data-driven techniques, and mathematical modeling in his work. He has published one book, over 35 peer-reviewed publications, and multiple conference papers. His work has been funded by several private and government agencies including Oculus, FHWA, and FDOT. He is a senior member of IEEE and serves as an associate editor of IEEE Transactions on Intelligent Transportation Systems. \\
Email: \url{shaurya.agarwal@ucf.edu} \\
Phone: +1 (407) 823-6205
\end{IEEEbiography}

\vfill

\end{document}